\documentclass[10pt,twocolumn,letterpaper]{article}

\usepackage[pagenumbers]{cvpr} 

\definecolor{cvprblue}{rgb}{0.21,0.49,0.74}
\usepackage[pagebackref,breaklinks,colorlinks,allcolors=cvprblue]{hyperref}

\usepackage{tabu}                      
\usepackage{booktabs}                  
\usepackage{lipsum}                    
\usepackage{mwe}                       

\usepackage{mathptmx}                  

\usepackage{amsmath}
\usepackage{amssymb}   
\usepackage{multirow}                  
\usepackage{xcolor}
\usepackage[normalem]{ulem} 

\def\paperID{*****} 
\def\confName{CVPR}
\def\confYear{2026}

\title{RSATalker: Realistic Socially-Aware Talking Head Generation for Multi-Turn Conversation}

\author{
    {\normalsize Peng Chen$^{1,2}$ \quad Xiaobao Wei$^{1,2}$ \quad Yi Yang$^{1}$ \quad Naiming Yao$^{1}$ \quad Hui Chen$^{1,2,\dagger}$ \quad Feng Tian$^{1,2}$}\\
    {\normalsize $^{1}$Institute of Software, Chinese Academy of Sciences}\\
    {\normalsize $^{2}$University of Chinese Academy of Sciences}\\
}

\begin{document}
\maketitle

\begin{abstract}
Talking head generation is increasingly important in virtual reality (VR), especially for social scenarios involving multi-turn conversation. Existing approaches face notable limitations: mesh-based 3D methods can model dual-person dialogue but lack realistic textures, large-model-based 2D methods produce natural appearances but incur prohibitive computational costs. Recently, 3D Gaussian Splatting (3DGS)-based methods achieve efficient and realistic rendering but remain speaker-only and ignore social relationships. We introduce RSATalker, the first framework that leverages 3DGS for realistic and socially-aware talking head generation, with support for multi-turn conversation. Our method first drives mesh-based 3D facial motion from speech, then binds 3D Gaussians to mesh facets to render high-fidelity 2D avatar videos. To capture interpersonal dynamics, we propose a socially-aware module that encodes social relationships, including blood and non-blood as well as equal and un-equal, into high-level embeddings through a learnable query mechanism. We design a three-stage training paradigm and construct the RSATalker dataset with speech–mesh–image triplets annotated with social relationships. Extensive experiments demonstrate that RSATalker achieves state-of-the-art performance in both realism and social awareness. The code and dataset will be released. 
\end{abstract}
\renewcommand{\thefootnote}{\fnsymbol{footnote}} 
\footnotetext[2]{Corresponding author.}
\section{Introduction}

\begin{figure*}[t]
    \centering
    \includegraphics[width=1.0\textwidth]{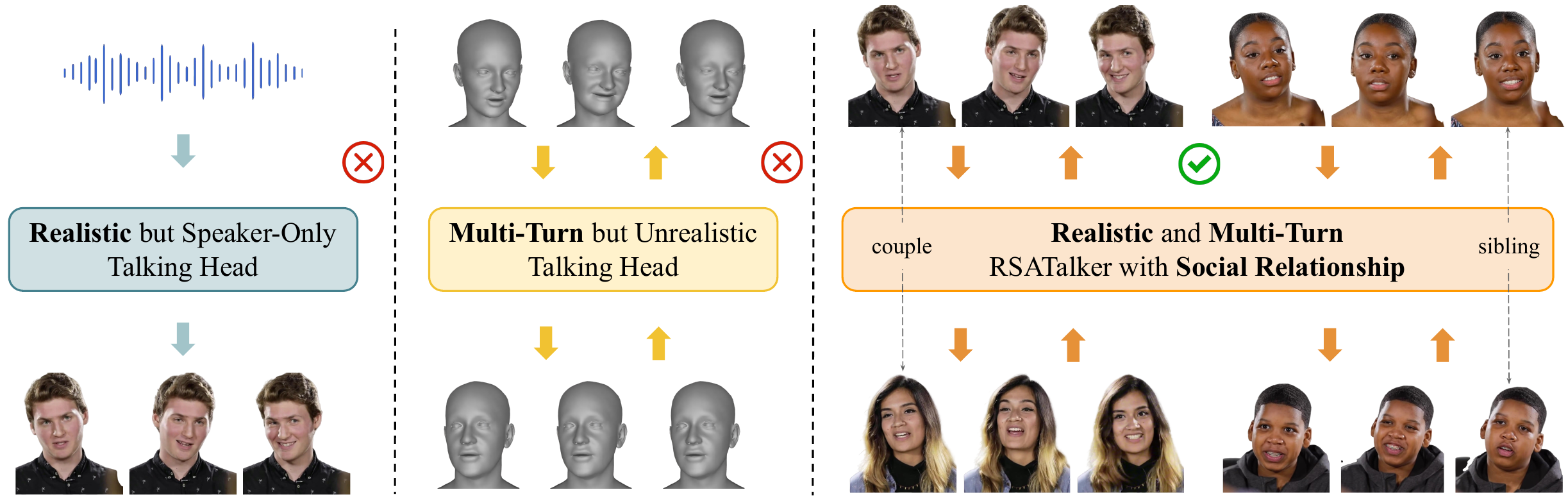}
    \caption{\textbf{Pipeline of RSATalker.}
    We introduce RSATalker, a novel framework that leverages 3D Gaussian Splatting (3DGS) to generate highly realistic and socially-aware talking heads, specifically designed for dynamic speaker–listener interaction scenarios.
    }
    \label{fig:pipeline}
\end{figure*}


In recent years, the advancement of deep learning techniques and computer graphics technologies has significantly promoted the development of talking head generation~\cite{bai2024bring, chen2025diffusiontalker, christoff2023application}. This capability has gained considerable importance in applications within virtual reality (VR)~\cite{liang20242024vrondigital, yang2024digitalization}, especially in immersive social interaction scenarios that involve multi-turn conversation between two participants~\cite{peng2025dualtalk, Learning-to-Listen}. In such settings, the roles of speaker and listener alternate dynamically over the course of dialogue. The outputs of talking head generation systems can generally be categorized into two representational paradigms: those based on 3D meshes and those relying on 2D imagery.


For 3D mesh-based methods, most approaches generate vertices or blendshapes~\cite{lewis2014practice} using Transformer- or Diffusion-based models to drive head mesh models such as FLAME~\cite{li2017learning}, thereby enabling talking head animations. Representative works include FaceFormer~\cite{fan2022faceformer}, EmoTalk~\cite{peng2023emotalk}, DiffusionTalker~\cite{chen2025diffusiontalker}, and Learning to Listen. However, these methods are typically ``speaker-only'' or ``listener-only.'' For multi-turn conversation, DualTalk~\cite{peng2025dualtalk} made an initial attempt by modeling dual-person dialogues, which marked an important step forward. 
Nevertheless, these approaches struggle to achieve photorealistic textures, accurate color reproduction, and naturalistic dynamic behaviors (e.g., facial expressions or micro-movements), frequently inducing the `uncanny valley'~\cite{schwind2018avoiding, katsyri2015review} effect in VR environments due to perceptible inconsistencies that trigger visceral discomfort.

In contrast, 2D image-based methods address this limitation by driving RGB human faces to generate photorealistic talking heads. Current approaches can be broadly grouped into two paradigms. First, large-scale pretrained models based on Stable Diffusion~\cite{rombach2022high} or Diffusion Transformer (DIT)~\cite{peebles2023scalable}, such as Wan~\cite{wan2025} and HunyuanVideo-Avatar~\cite{chen2025hunyuanvideo}, exhibit strong generalization capabilities, but their massive parameter sizes lead to extremely high computational costs, rendering them impractical for VR deployment. Second, Neural Radiance Field~\cite{mildenhall2021nerf} (NeRF)- and 3D Gaussian Splatting~\cite{3dgs} (3DGS)-based methods, such as ER-NeRF~\cite{er-nerf}, SyncTalk~\cite{peng2024synctalk}, and GaussianTalker~\cite{cho2024gaussiantalker}, achieve high rendering efficiency and accuracy at relatively low computational costs, making them well-suited for VR applications. However, existing approaches remain fundamentally ``speaker-only,'' without modeling the dynamics of multi-turn conversation. 

Moreover, in the context of social interactions during multi-turn conversation, the social relationship between two speakers plays a decisive role in shaping facial expressions, head movements, and other nonverbal behaviors, a crucial factor that previous works have not taken into account~\cite{frith2009role, pelachaud1996generating}. In real multi-turn conversation, these relationships strongly influence both expressiveness and responsiveness: for instance, the conversation between close friends often involve relaxed expressions, frequent smiling, and dynamic gestures, whereas professional dialogues between a superior and a subordinate tend to be more restrained, with controlled gaze shifts and formal expressions~\cite{rafaeli1989expression, tannen2005conversational}. Such interpersonal dynamics accumulate over multiple dialogue rounds, directly affecting how smoothly roles transition between speaker and listener, and determining the overall naturalness of the interaction. Ignoring this relationship leads to talking heads that may appear visually plausible but socially inconsistent, ultimately limiting their effectiveness in VR-based social communication.

In summary, as shown in Fig.~\ref{fig:teaser}, current methods for talking head generation in multi-turn conversation face the following limitations: 
(1) Existing realistic talking head generation methods only focus on the speaker-only setting.
(2) Mesh-based 3D approaches can achieve dynamic multi-turn interaction between speakers and listeners, but the results are unrealistic.
(3) Existing talking head methods overlook the importance of social relationships in multi-turn conversation.


\begin{figure*}[t]
    \centering
    \includegraphics[width=1.0\textwidth]{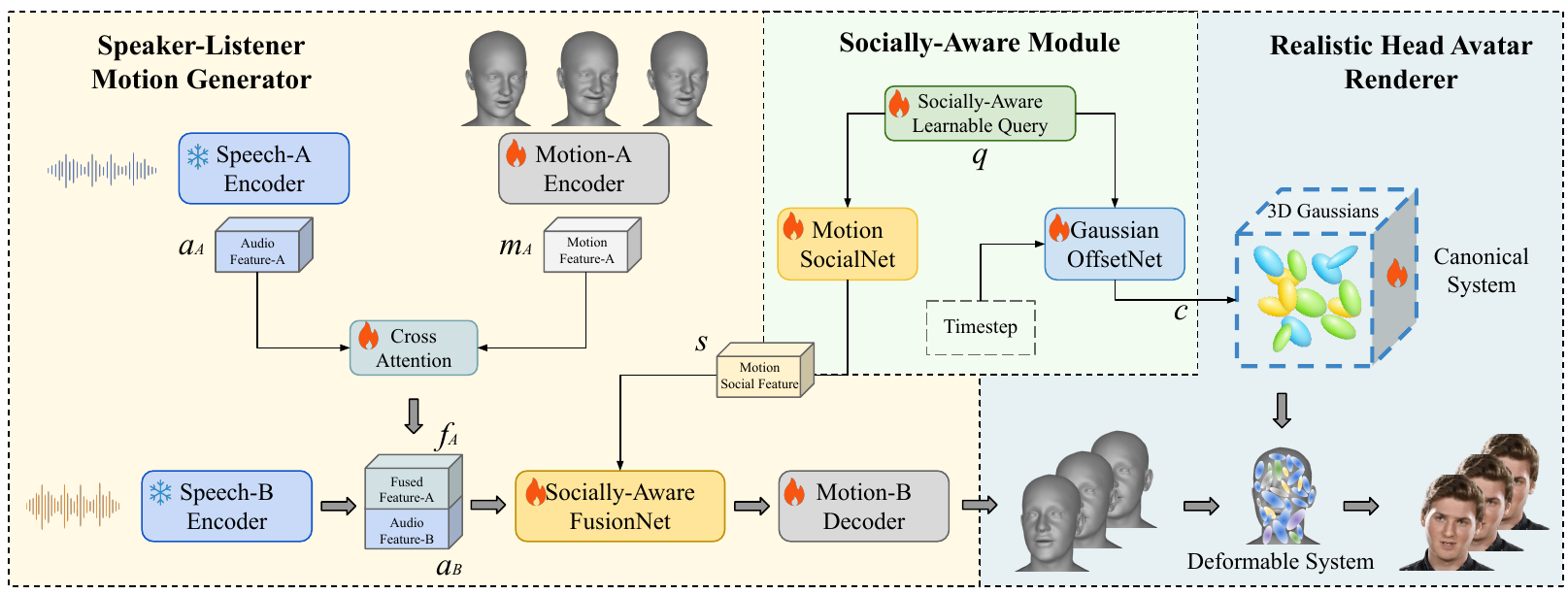}
    \caption{\textbf{Pipeline of RSATalker.}
    We introduce RSATalker, a novel framework that leverages 3D Gaussian Splatting (3DGS) to generate highly realistic and socially-aware talking heads, specifically designed for dynamic speaker–listener interaction scenarios.
    }
    \label{fig:pipeline}
\end{figure*}

To address the above issues, we propose RSATalker, the first method that leverages 3DGS to achieve \textbf{R}ealistic and \textbf{S}ocially-\textbf{A}ware talking head generation between speakers and listeners. 
Our approach combines 3D mesh with 3DGS and consists of three modules: a speaker–listener motion generator, a realistic head avatar renderer, and a socially-aware module. The speaker–listener motion generator is built upon 3D mesh and integrates multiple modalities such as visual and audio cues to infer smooth facial motion.
Subsequently, the realistic head avatar renderer associates 3D Gaussian primitives with the mesh’s triangular facets and exploits mesh deformations to produce high-fidelity 2D avatar videos. 
To incorporate the social relationship between two people during the conversation, we introduce a socially-aware module. Specifically, we categorize social relationships along two dimensions—\emph{blood vs. non-blood} and \emph{equal vs. non-equal}—and employ a learnable query mechanism to abstract these relationships into high-level embeddings, which are integrated into the model features. 
We adopt a three-stage training paradigm, consisting of two cold-start stages followed by an end-to-end training stage. Moreover, we construct the RSATalker dataset, which contains speech–mesh–image triplets of different social relationships. Extensive experiments demonstrate that our method achieves state-of-the-art performance in this field.

In summary, the main contributions of this work are as follows:
\begin{itemize}
\item We propose RSATalker, which leverages 3DGS to achieve realistic and socially-aware talking head generation for speaker–listener interaction.
\item We introduce a socially-aware module that incorporates the crucial interpersonal relationship between conversational partners to investigate its impact on talking head generation.
\item We construct the RSATalker dataset, a realistic multi-turn conversation dataset focusing on social relationships, and design a three-stage training strategy to optimize the performance of talking head generation.
\end{itemize}

\section{Related Work}
\subsection{Realistic Speech-Driven Talking Head Generation}
Recent advances in talking head generation have been largely driven by large-scale pretrained models, which achieve impressive photorealism. 
MEMO~\cite{li2025memo} introduces a memory-guided diffusion framework that enhances long-term identity consistency and emotion-aware expression alignment through temporal modeling and audio-driven emotion modulation. 
Sonic~\cite{ji2025sonic} improves photorealism and expressiveness by incorporating audio-aligned identity units and semantic motion decomposition into a coarse-to-fine denoising pipeline, achieving accurate lip-sync and strong identity preservation across diverse speakers.
EmotiveTalk~\cite{wang2025emotivetalk} disentangles emotion and content in both audio and visual domains, allowing fine-grained control over identity, emotion intensity, and lip synchronization through an emotion-aware conditioning mechanism.
HunyuanVideo-Avatar~\cite{chen2025hunyuanvideo} employs a unified diffusion model trained on large-scale video data to synthesize personalized avatars with diverse expressions and head poses, integrating identity encoding and controllable condition inputs for high-fidelity video generation. 
To address efficiency, VASA-1~\cite{xu2024vasa} proposes a real-time pipeline that leverages a latent face embedding and controllable keypoints to produce realistic, expressive talking heads from a single image and audio, offering a favorable speed–quality trade-off.

While recent diffusion-based approaches have achieved impressive realism, their reliance on large-scale pretrained models often results in high computational cost and slow inference, limiting their deployment in interactive or real-time scenarios. 
To address these limitations, a line of research has turned to neural scene representation methods, particularly those based on Neural Radiance Field (NeRF)~\cite{mildenhall2021nerf, wei2024nto3d, hong2022headnerf, guo2021ad} and 3D Gaussian Splatting (3DGS)~\cite{3dgs, wei2025gazegaussian, wei2025graphavatar, wei2025emd}, which offer significantly faster rendering while maintaining high visual quality.
ER-NeRF~\cite{er-nerf} extends neural radiance fields with expression-aware radiance modeling and audio-driven deformation fields, enabling high-fidelity rendering of dynamic facial expressions and lip-synchronized speech from monocular input, along with free-viewpoint control.
To further improve efficiency, SyncTalk~\cite{peng2024synctalk} adopts a structured implicit representation combining audio-driven keypoint control and neural blendshapes, effectively decoupling identity, expression, and pose for real-time, expressive NeRF-based talking head generation.
Moving beyond radiance fields, GaussianTalker~\cite{cho2024gaussiantalker} pioneers the use of 3D Gaussian Splatting for talking head synthesis by learning a dynamic neural blendshape basis over Gaussians, achieving real-time rendering with accurate lip sync and high identity fidelity.
TalkingGaussian~\cite{li2024talkinggaussian} generalizes this idea by introducing a unified 3DGS framework that incorporates audio-driven dynamic Gaussians with decomposed appearance and pose control, enabling disentangled and expressive talking head generation with high efficiency.
Building on this, EmoTalkingGaussian~\cite{cha2025emotalkinggaussian} further enhances the expressiveness of 3DGS-based models by introducing emotion-aware dynamic Gaussians and semantic priors, allowing fine-grained. 

Despite their advantages in rendering speed and fidelity, these NeRF- and 3DGS-based approaches are typically designed for single-speaker scenarios and lack mechanisms to model dual-speaker interaction, such as turn-taking and socially-aware behavioral cues between interlocutors. 

\subsection{Multi-Turn Conversation}

Modeling multi-turn conversation in talking head generation remains a largely underexplored yet crucial area. 
Early attempts, such as RealTalk~\cite{ji2024realtalk} and Learning to Listen~\cite{Learning-to-Listen}, primarily focus on the listener-only setting. These approaches generate short-term nonverbal feedback---such as nodding, blinking, or subtle head tilts---in response to the speaker’s audio-visual cues. While such methods improve realism in passive listening scenarios, they are inherently limited: they cannot model role switching between speakers, nor sustain coherent bidirectional interactions across multiple turns of dialogue. Consequently, they fail to capture the dynamic reciprocity that defines natural human conversation.

To move beyond passive listening, DualTalk~\cite{peng2025dualtalk} introduces the first multi-turn conversation framework for talking head generation. By jointly modeling both conversational roles, DualTalk enables more interactive behaviors, including role alternation, mutual gaze control, and motion-aware feedback. This represents an important step toward more natural multi-turn dialogue modeling. However, despite these advances, DualTalk is still built upon relatively coarse 3D mesh representations. As a result, it lacks the ability to synthesize realistic textures, fine-grained appearance details, or natural color variations, which are essential for achieving photorealistic quality. This limitation significantly reduces its applicability in VR and other immersive environments, where users expect both high visual fidelity and socially consistent interaction. 

In addition, prior studies have primarily focused on incorporating emotion and personalization into talking head generation~\cite{peng2023emotalk, chen2025diffusiontalker}, but have not explored the impact of social relationships in multi-turn conversation.

In summary, prior works on multi-turn conversation either restrict themselves to short-term listener feedback or adopt coarse geometric models that cannot deliver photorealistic visual quality, while overlooking this critical factor of social relationships. Bridging this gap requires a new generation of models capable of jointly modeling dynamic role switching, rich nonverbal behaviors, and high-fidelity appearance, all within a socially aware conversational framework.



\section{Method}

\subsection{Preliminaries}
In 3D Gaussian Splatting~\cite{3dgs} (3DGS), a 3D scene is represented as a collection of Gaussian primitives. Each primitive is parameterized by its center position $\mathbf{p_k} \in \mathbb{R}^3$ and a covariance matrix $\Sigma \in \mathbb{R}^{3 \times 3}$. The Gaussian density at a spatial point $\mathbf{p}$ can be expressed as:

\begin{equation}
G(\mathbf{p}) = \exp\left(-\tfrac{1}{2} (\mathbf{p} - \mathbf{p_k})^\top \Sigma^{-1} (\mathbf{p} - \mathbf{p_k})\right)
\end{equation}

To make the covariance easier to interpret and manipulate, it is commonly decomposed into a rotation matrix $R$ and a scaling matrix $S$, such that:

\begin{equation}
\Sigma = R S S^\top R^\top
\end{equation}
where $R$ encodes the orientation of the Gaussian in 3D space, while $S$ determines its anisotropic spread along different axes.

During the rendering stage, these 3D Gaussian primitives are projected onto the image plane. This is achieved by applying the world-to-camera transformation matrix $W$, which maps points from world coordinates to camera coordinates, and a local affine transformation matrix $J$, which adjusts the shape of the Gaussian under perspective projection. The result of this projection is a set of 2D Gaussian primitives, each with a covariance matrix $\Sigma^{2D}$. The corresponding projected Gaussian is denoted as $G^{3D}$ when expressed in screen space.
Finally, image formation is carried out via alpha blending. In this process, the contribution of each projected Gaussian is accumulated along the viewing direction. The color at a given pixel $\mathbf{x}$ is computed as:

\begin{equation} \label{eq:splatting}
\mathbf{c}(\mathbf{x}) = \sum_{k=1}^{K} \mathbf{c_k} \alpha_k G^{3D}_k(\mathbf{x}) \prod_{j=1}^{k-1} \big(1 - \alpha_j G^{3D}_j(\mathbf{x})\big)
\end{equation}
where $\mathbf{c_k}$ denotes the color attribute of the $k$-th Gaussian primitive, and $\alpha_k$ represents its opacity. 

\subsection{Overview}
RSATalker consists of three modules: speaker-listener motion generator, realistic head avatar renderer, and socially-aware module.

For the speaker-listener motion generator, the inputs are speaker-A's speech and FLAME-based~\cite{li2017learning} motion parameters, as well as speaker-B's speech. 
Through a cross-attention mechanism, a socially-aware fusionnet, and a motion decoder, it predicts speaker-B's facial motion, which can drive the FLAME model to generate 3D head avatar facial animations.  
The realistic head avatar renderer binds 3D Gaussians to the corresponding triangular facets of the FLAME model. As the deformable FLAME model moves, the 3D Gaussians are indirectly driven to move. For a given timestep, the 3D Gaussians are rendered into an RGB image through Gaussian splatting. Multiple timesteps collectively render a talking head video.  
The socially-aware module employs learnable queries to process learnable social embeddings through the motion socialnet and Gaussian offsetnet, respectively producing motion social features and Gaussian offset corrections. This injects social relationship information into the first two modules.

\subsection{Speaker-Listener Motion Generator}

The primary function of this module is to capture the relationship between speech and facial motion. It conditions on the audiovisual information of the interlocutor (i.e., speaker-A) to better guide the model in generating the facial motion of the current speaker (i.e., speaker-B) within social interaction scenarios.

Specifically, the speech signals from speaker-A and speaker-B ($\text{speech}_A$, $\text{speech}_B$) are processed by a pretrained speech encoder Wav2Vec 2.0 to obtain their corresponding audio features $a_A$ and $a_B$. Meanwhile, the facial motion of speaker-A, represented by FLAME parameters, is passed through a motion encoder to yield the motion feature $m_A$.
To effectively capture speaker-A’s audiovisual information, we fuse $a_A$ and $m_A$ using a cross-attention mechanism. The fusion process is formulated as:

\begin{equation}
f_A = \text{CrossAttn}\big(\mathbf{Q}=a_A,\; \mathbf{K}=m_A,\; \mathbf{V}=m_A \big)
\end{equation}
where $f_A$ denotes the fused feature of speaker-A.

\begin{figure}[t]
    \centering
    \includegraphics[width=0.5\textwidth]{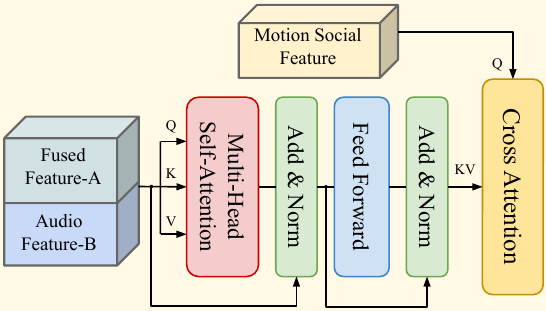}
    \caption{The structure of the socially-aware fusion network.}
    \label{fig:fusion}
\end{figure}

Subsequently, we concatenate $f_A$ and $a_B$, and feed the combined features into a socially-aware fusion network, which injects social relationship information into the motion representation. 
As illustrated in Fig.~\ref{fig:fusion}, the concatenated features $[f_A;a_B]$ are first processed by a Transformer encoder to model their temporal dependencies and contextual interactions. 
The encoded representation is then used as the keys and values in a cross-attention mechanism, where the query is the motion social feature $s$ generated from the socially-aware module:  

\begin{equation}
z = \text{CrossAttn}\big(\mathbf{Q}=s,\; \mathbf{K}=[f_A;a_B],\; \mathbf{V}=[f_A;a_B] \big),
\end{equation}
where $[\cdot;\cdot]$ denotes the feature concatenation operation.  
This design allows the query of motion social feature $s$ to selectively attend to the fused motion–audio features according to the underlying social relationship, thereby embedding socially conditioned dynamics into the motion representation.  

Finally, the socially-aware motion representation $z$ is passed through a Transformer-based motion decoder, which generates the target facial motion sequence of speaker B. The predicted motion is directly mapped onto the FLAME parametric head model, enabling the rendering of high-fidelity and socially consistent facial dynamics for the multi-turn conversation.




\subsection{Realistic Head Avatar Renderer}

To generate realistic talking heads, we adopt a coordinate system conversion strategy inspired by the 3D Gaussian rigging approach in GaussianAvatars~\cite{qian2024gaussianavatars}. The key idea is to establish a mapping between the FLAME mesh and the 3D Gaussians, where the mapping essentially encodes the transformation between the canonical coordinate system and the deformable coordinate system.

At initialization, each triangular facet of the FLAME mesh is assigned a corresponding 3D Gaussian. If the FLAME mesh has $N$ triangular facets, then $N$ 3D Gaussians are initialized, which we refer to as the anchor Gaussians.
In the canonical coordinate system, the Gaussian parameters are set to default states: the position $\mu_c$ is initialized to zero, the rotation matrix $r_c$ is the identity matrix, the scaling $s_c$ is a unit vector, and the opacity $\alpha$ is set to zero. In contrast, in the deformable coordinate system, these Gaussian parameters are determined by the geometry of the associated mesh triangle. Through this mapping, the 3D Gaussians are transformed from the canonical system to the deformable one, enabling the Gaussians to follow mesh deformations.

\begin{figure}[t]
    \centering
    \includegraphics[width=0.3\textwidth]{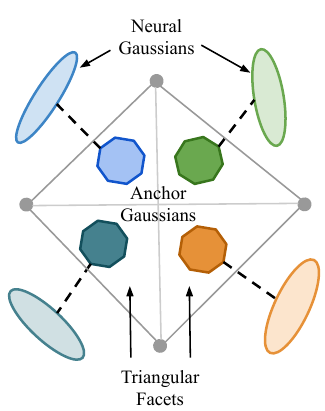}
    \caption{The anchor–neural structure of the 3D Gaussians.}
    \label{fig:gaussian}
\end{figure}

During training, 3DGS performs adaptive density control, under which existing anchor Gaussians may be cloned and split. The newly cloned and split Gaussians are referred to as the neural Gaussians. Each neural Gaussian is bound to its anchor Gaussian, as illustrated in Fig.~\ref{fig:gaussian}, thereby forming an anchor-neural relationship.

In the deformable coordinate system, the parameters of a 3D Gaussian are defined as the position $\mu_d$, rotation matrix $r_d$, scaling $s_d$, and opacity $\alpha$. The transformation of Gaussian rotation and scaling is formulated as:
\begin{equation}
r_d = R r_c, \quad s_d = \lambda s_c
\end{equation}
where $R$ is a rotation matrix constructed from the triangle’s edge direction, its normal vector, and their cross-product. The factor $\lambda$ corresponds to the average length of an edge and its perpendicular, serving as a global scale factor for both position and size.

The position transformation between the mesh and the 3D Gaussian is defined as:
\begin{equation}
\mu_d = \lambda R \mu_c + T + c
\end{equation}
where $T$ denotes the centroid of the corresponding triangle in global space. To reduce alignment errors, we further introduce Gaussian offset corrections $c$ (with $N$ dimensions, corresponding to the $N$ anchor Gaussians) from the socially-aware module, which adaptively corrects positional deviations over time. Each neural Gaussian shares the same Gaussian offset correction with its anchor Gaussian.

\begin{figure}[t]
    \centering
    \includegraphics[width=0.4\textwidth]{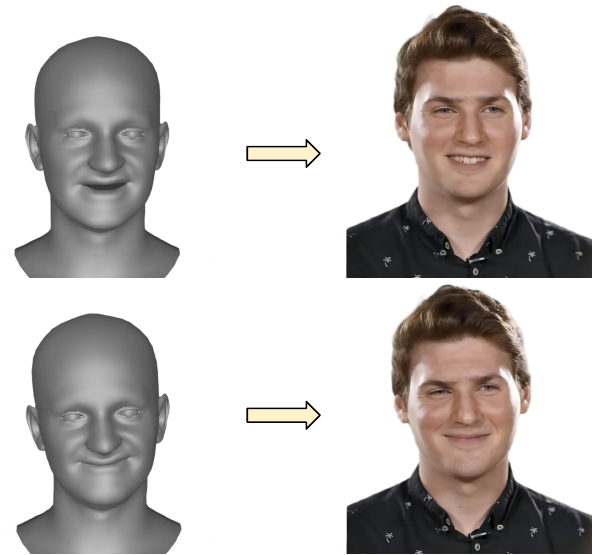}
    \caption{In the deformable system, the motion of the mesh can drive the 3DGS to render a realistic 2D talking head.}
    \label{fig:mesh_gs}
\end{figure}

At each timestep, all 3D Gaussians in the deformable system are rendered into an image through splatting as formulated in Eq.~\ref{eq:splatting}. The images generated across all timesteps can then be composed into a realistic head avatar video, as shown in Fig.~\ref{fig:mesh_gs}.

\subsection{Socially-Aware Module}

We categorize social relationships in multi-turn conversation along two orthogonal dimensions: \emph{blood vs. non-blood} and \emph{equal vs. non-equal}, which yield four possible combinations of interpersonal contexts. 
For example, a young couple falls under the category of non-blood and equal, a mother-son relationship corresponds to blood and non-equal, while an interaction between superior–subordinate colleagues is classified as non-blood and non-equal. This taxonomy provides a compact yet expressive representation of the diverse social dynamics that naturally arise in human dialogues.

To model these relationships, we maintain two sets of embeddings, each encoding one dimension of the social taxonomy. Specifically, each set consists of two orthogonal embeddings, allowing us to represent the binary states in a disentangled manner. Given an input conversation sample, the embedding corresponding to the detected social relationship is retrieved and fused to form a structured query $q$ of the interpersonal context. This query is then processed by a dedicated motion socialnet, implemented as a two-layer MLP, to generate the motion social feature $s$. 
The feature $s$ serves as a high-level relational prior that modulates the facial motion dynamics through cross-attention with the motion modality. 
In parallel, we introduce a Gaussian offsetnet, which is also a two-layer MLP. It takes as input $q$ and the timestep $t$, and generates Gaussian offset corrections $c$. The formulation is as follows:

\begin{equation}
c = \text{GON}(q, t)
\end{equation}
where GON denotes the Gaussian offsetnet.
These corrections are applied during the coordinate system conversion step, adaptively adjusting the positional parameters of the 3D Gaussian primitives. In this way, the socially conditioned information not only influences temporal motion generation but also fine-tunes the geometric rendering of facial details.

Importantly, all parameters in this socially-aware module are learnable, enabling the system to jointly optimize motion dynamics and appearance adaptation under different social contexts. This design allows the model to flexibly capture subtle interpersonal variations—such as restrained facial expressions in the hierarchical conversation or expressive cues in intimate interactions—ultimately leading to more socially consistent talking head generation.

\subsection{Three-Stage Training}

We adopt a three-stage training paradigm to progressively build up the proposed framework. In the first stage, we pretrain the speaker–listener motion generator through a cold start, where a large-scale dataset is used to learn the general interaction patterns of multi-turn conversation. The training objective is defined as
\begin{equation}
L_{\text{mesh}} = \| M - M_{\text{gt}} \|_2
\end{equation}
where $M$ denotes the facial motion parameterized by the FLAME model (e.g., expression and pose coefficients), and $M_{\text{gt}}$ is the corresponding ground truth.

In the second stage, we pretrain the realistic head avatar renderer by reconstructing the 3DGS representation of the speaker. This step allows the model to acquire the ability to render speaker-specific head and facial motion. We calculate the $L_1$ loss between the rendered images and the ground truth, including a D-SSIM term:
\begin{equation}
L_{\text{image}} = (1 - \lambda)L_1 + \lambda L_{\text{D-SSIM}}
\end{equation}
where $\lambda$ is set to 0.2. To ensure that the locations of the 3D Gaussians and the corresponding underlying mesh remain adequately aligned in the deformable system, we constrain the canonical position $\mu_c$ using $L_{\text{pos}}$:
\begin{equation}
L_{\text{pos}} = \left\| \max \left( \mu_c, \epsilon_{\text{pos}} \right) \right\|_2
\end{equation}
Similarly, for the Gaussian offset corrections $c$, we also restrict their magnitude:
\begin{equation}
L_{\text{offset}} = \left\| \max \left( c, \epsilon_{\text{offset}} \right) \right\|_2
\end{equation}
where the thresholds are set as $\epsilon_{\text{pos}} = \epsilon_{\text{offset}} = 1$.

Finally, in the third stage, we introduce the socially-aware module and perform end-to-end joint training, which integrates social relationship embeddings into the overall system. The joint optimization objective is formulated as
\begin{equation}
L_{\text{joint}} = L_{\text{mesh}} + \lambda_1 L_{\text{image}} + \lambda_2 L_{\text{pos}} + \lambda_3 L_{\text{offset}}
\end{equation}
where $\lambda_1$ is set to 0.5, and $\lambda_2 = \lambda_3 = 0.01$. Except for the parameters of the speech encoder, all other model parameters are trained.

\section{Experiments}

\begin{table*}[t]
\centering
\caption{Quantitative evaluation on the test set of DualTalk dataset.}
\resizebox{0.85\linewidth}{!}{
\begin{tabular}{@{}lccccccccc@{}}
\toprule
& \multicolumn{3}{c}{\textbf{FD} $\downarrow$} 
& \multicolumn{3}{c}{\textbf{P-FD} $\downarrow$} 
& \multicolumn{3}{c}{\textbf{MSE} $\downarrow$} \\ 
\cmidrule(lr){2-4} \cmidrule(lr){5-7} \cmidrule(lr){8-10}
\textbf{Methods} 
& \textbf{EXP} & \textbf{JAW} & \textbf{POSE} 
& \textbf{EXP} & \textbf{JAW} & \textbf{POSE} 
& \textbf{EXP} & \textbf{JAW} & \textbf{POSE} \\ 
&  & $\times 10^3$ & $\times 10^2$ 
&  & $\times 10^3$ & $\times 10^2$ 
& $\times 10^1$ & $\times 10^3$ & $\times 10^2$ \\ 
\midrule
FaceFormer~\cite{fan2022faceformer} 
& 34.90 & 5.40 & 8.00 
& 34.90 & 5.40 & 8.00 
& 6.97 & 1.80 & 2.67 \\
CodeTalker~\cite{xing2023codetalker} 
& 48.57 & 6.89 & 10.74 
& 48.57 & 6.89 & 10.74 
& 9.71 & 2.29 & 3.58 \\
EmoTalk~\cite{peng2023emotalk} 
& 29.86 & 4.33 & 7.54 
& 30.20 & 4.36 & 7.58 
& 6.88 & 1.76 & 2.59 \\
DiffusionTalker~\cite{chen2025diffusiontalker}
& 30.67 & 4.89 & 8.23 
& 36.71 & 5.98 & 7.43 
& 6.54 & 1.63 & 2.57 \\
SelfTalk~\cite{peng2023selftalk} 
& 35.77 & 5.49 & 8.14 
& 35.77 & 5.49 & 8.14 
& 7.15 & 1.83 & 2.71 \\
L2L~\cite{Learning-to-Listen} 
& 24.61 & 3.69 & 7.08 
& 24.99 & 3.74 & 7.13 
& 5.68 & 1.48 & 2.49 \\
DualTalk~\cite{peng2025dualtalk}
& 11.14 & \textbf{1.90} & 3.83 
& 11.88 & 1.97 & \textbf{3.97} 
& 3.59 & \textbf{1.04} & 1.72 \\
RSATalker
& \textbf{10.68} & 2.05 & \textbf{3.43} 
& \textbf{10.21} & \textbf{1.82} & 4.34 
& \textbf{2.98} & 1.12 & \textbf{1.65} \\
\bottomrule
\end{tabular}}
\label{tab:quan_mesh}
\end{table*}

\subsection{RSATalker Dataset Construction}
In multi-turn conversational talking head scenarios, existing datasets often lack one or more modalities. For example, RealTalk~\cite{ji2024realtalk} contains only speech and 2D video, whereas DualTalk~\cite{peng2025dualtalk} provides only speech and 3D mesh data. This limitation motivates the need for a dataset that includes all three modalities—speech, 3D mesh, and video—to fully capture facial movements in multi-turn interactions. To address this, we construct a speech–mesh–video triplet dataset by combining information from RealTalk and DualTalk, providing a more comprehensive resource for learning realistic talking head dynamics.
Specifically, we first leverage two social relationship dimensions in DualTalk (\emph{blood/non-blood} and \emph{equal/non-equal}) to form four typical relationship scenarios: blood \& equal (e.g., sibling), blood \& non-equal (e.g., mother and son), non-blood \& equal (e.g., couple), and non-blood \& non-equal (e.g., superior–subordinate colleagues). We then retrieve the corresponding video segments from RealTalk using the speech clips from these data, precisely localizing their temporal positions via audio feature matching and cropping the corresponding video. This results in a complete speech–mesh–video triplet dataset, providing richer support for multi-modal modeling.

\subsection{Experimental Settings}
\subsubsection{Baselines}
To comprehensively evaluate the capability of RSATalker, we selected two types of talking head methods, namely mesh-based and image-based approaches, as baselines. 
The former is evaluated on the test set of DualTalk dataset, while the latter is assessed on the test set of RSATalker dataset.
The mesh generation methods include FaceFormer~\cite{fan2022faceformer}, CodeTalker~\cite{xing2023codetalker}, DiffusionTalker~\cite{chen2025diffusiontalker}, EmoTalk~\cite{peng2023emotalk}, SelfTalk~\cite{peng2023selftalk}, Learning-to-Listen(L2L)~\cite{Learning-to-Listen}, and DualTalk~\cite{peng2025dualtalk}. 
The image generation methods include ER-NeRF~\cite{li2023efficient}, SyncTalk~\cite{peng2024synctalk}, and GaussianTalker~\cite{cho2024gaussiantalker}. 
For speaker-only and listener-only methods, we could only use the speech of one single speaker as input.

\subsubsection{Metrics}
We employ two sets of quantitative metrics to evaluate the mesh and image, one set of qualitative talking head image visualizations, and one socially-aware user study. 

For mesh-based methods, we use metrics such as Frechet Distance (FD), Paired Frechet Distance (P-FD), and Mean Squared Error (MSE) to assess the accuracy and smoothness of facial motions.

For the image–based methods, we adopt L1, PSNR, SSIM, and LPIPS to measure the quality of the generated images and the realism of the talking heads. Detailed experimental results can be found in the quantitative evaluation section. 

In addition, to assess the performance of talking head generation in capturing social relationships, we propose a novel metric termed ``socially-aware user study''. This metric relies on human evaluation, where professional annotators assign scores to all methods. The specific experimental setup and results are presented in the qualitative evaluation section.




\subsection{Quantitative Evaluation}
\subsubsection{Mesh-based methods}

For mesh-based methods, the results are shown in Tab.~\ref{tab:quan_mesh}. We report three metrics—FD, P-FD, and MSE—each further evaluated across three dimensions: expression, jaw, and head pose. All models were tested on the DualTalk dataset test set.

Among them, FaceFormer, CodeTalker, EmoTalk, DiffusionTalker, and SelfTalk are speaker-only methods. Suppose the two participants are speaker A and speaker B: in this setting, these methods can only take speaker B’s speech as input, without conditioning on speaker A’s facial motion or speech.
However, in multi-turn dialogue scenarios, speaker B may act as a listener. In such cases, speaker B’s speech often consists only of filler tokens (e.g., “uh,” “hmm,” “mm”), which lack meaningful semantic content. If the model relies solely on speaker B’s speech input, it struggles to generate natural facial expressions and head poses, leading to higher errors across the reported metrics.

L2L addresses this limitation by effectively capturing speaker B’s facial behaviors when acting as a listener, thereby reducing errors across all metrics. Nevertheless, it cannot handle multi-turn dialogues, which results in suboptimal performance when speaker B takes the role of the speaker. As a result, its overall performance still falls short of the best.

DualTalk, as the first method designed for multi-turn dialogue, introduces improvements in temporal modeling and feature fusion for both speakers and listeners. This significantly enhances the modeling of interactions between the two participants, yielding a notable performance gain over L2L. However, it does not consider the influence of social relationships in multi-turn conversation.

To overcome this limitation, RSATalker incorporates a socially-aware module, which classifies and extracts features based on different types of social relationships. This design further improves the accuracy of facial expressions and head poses in multi-turn dialogue scenarios.

\begin{table}[h]
    \centering
    \caption{Quantitative evaluation on the test set of RSATalker dataset.}
    \resizebox{0.50\textwidth}{!}{%
    \begin{tabular}{l|cccc} 
    \toprule
         Method & L1 $\downarrow$ & PSNR $\uparrow$ & SSIM $\uparrow$ &  LPIPS $\downarrow$  \\
         \midrule
         ER-NeRF~\cite{li2023efficient} & 0.0468 & 19.8098 & 0.8856 & 0.1413  \\
         SyncTalk~\cite{peng2024synctalk} & 0.0288 & 20.9312 & 0.9184 & 0.1035  \\
         GaussianTalker~\cite{cho2024gaussiantalker} & 0.0224 & 21.3262 & 0.9164 & 0.0967  \\
         RSATalker & \textbf{0.0198} & \textbf{22.9979} & \textbf{0.9377} & \textbf{0.0560}  \\
         \midrule

    \end{tabular}
    }
    \label{tab:quan_image}
\end{table}

\subsubsection{Image-based methods}

\begin{table*}[t]
    \centering
    \caption{User study results (0--100 scale). Each dimension is evaluated by three sub-criteria, and the final scores are averaged across all participants. RSATalker achieves the highest scores across all dimensions, with the most significant improvements observed in SRA. Each dimension mean is reported in the last column of the dimension.}
    \resizebox{\textwidth}{!}{%
    \begin{tabular}{l|cccc|cccc|cccc|c}
    \toprule
    \multirow{4}{*}{Method} 
        & \multicolumn{4}{c|}{Realism $\uparrow$} 
        & \multicolumn{4}{c|}{Fluency $\uparrow$} 
        & \multicolumn{4}{c|}{SRA $\uparrow$} 
        & \multirow{4}{*}{Total $\uparrow$} \\ 
    \cmidrule(lr){2-5} \cmidrule(lr){6-9} \cmidrule(lr){10-13}
        & \shortstack{Lip \\ Sync} 
        & \shortstack{Expr. \\ Naturalness} 
        & \shortstack{Visual \\ Realism} 
        & \shortstack{Mean}
        & \shortstack{Motion \\ Smoothness} 
        & \shortstack{Temporal \\ Consistency} 
        & \shortstack{Long-term \\ Stability} 
        & \shortstack{Mean}
        & \shortstack{Role \\ Consistency} 
        & \shortstack{Emotional \\ Tone} 
        & \shortstack{Interaction \\ Dynamics} 
        & \shortstack{Mean}
        &  \\
    \midrule
    ER-NeRF~\cite{li2023efficient} & 65.1 & 58.3 & 62.7 & 62.0 & 60.2 & 58.5 & 55.4 & 58.0 & 40.6 & 42.1 & 38.4 & 40.4 & 50.2 \\
    SyncTalk~\cite{peng2024synctalk} & 85.2 & 72.4 & 74.1 & 77.2 & 70.3 & 65.7 & 68.2 & 68.1 & 48.6 & 50.2 & 47.3 & 48.7 & 60.7 \\
    GaussianTalker~\cite{cho2024gaussiantalker} & 80.5 & 78.6 & 86.2 & 81.8 & 75.8 & 72.5 & 70.9 & 73.1 & 55.4 & 58.7 & 60.3 & 58.1 & 67.8 \\
    RSATalker & \textbf{90.7} & \textbf{86.4} & \textbf{88.2} & \textbf{88.4} & \textbf{88.5} & \textbf{86.1} & \textbf{85.6} & \textbf{86.7} & \textbf{84.7} & \textbf{82.3} & \textbf{86.5} & \textbf{84.5} & \textbf{86.1} \\
    \bottomrule
    \end{tabular}
    }
    \label{tab:user_study_full}
\end{table*}

2D image-based talking head methods better capture the realism of digital humans. We conducted the evaluation on the RSATalker dataset test set, with results shown in Tab.~\ref{tab:quan_image}. ER-NeRF and SyncTalk are NeRF-based methods, while GaussianTalker and RSATalker are based on 3DGS. Overall, compared with NeRF, 3DGS-based methods deliver higher rendering quality.

However, since ER-NeRF, SyncTalk, and GaussianTalker are all speaker-only methods, they cannot take speaker A’s facial motion and speech as conditional inputs and can only rely on speaker B’s speech, which leads to suboptimal rendering quality. By contrast, RSATalker benefits from its ability to support multi-turn dialogues and incorporate social relationship features, enabling it to significantly outperform the baselines.

\begin{figure*}[t]
    \centering
    \includegraphics[width=1.0\textwidth]{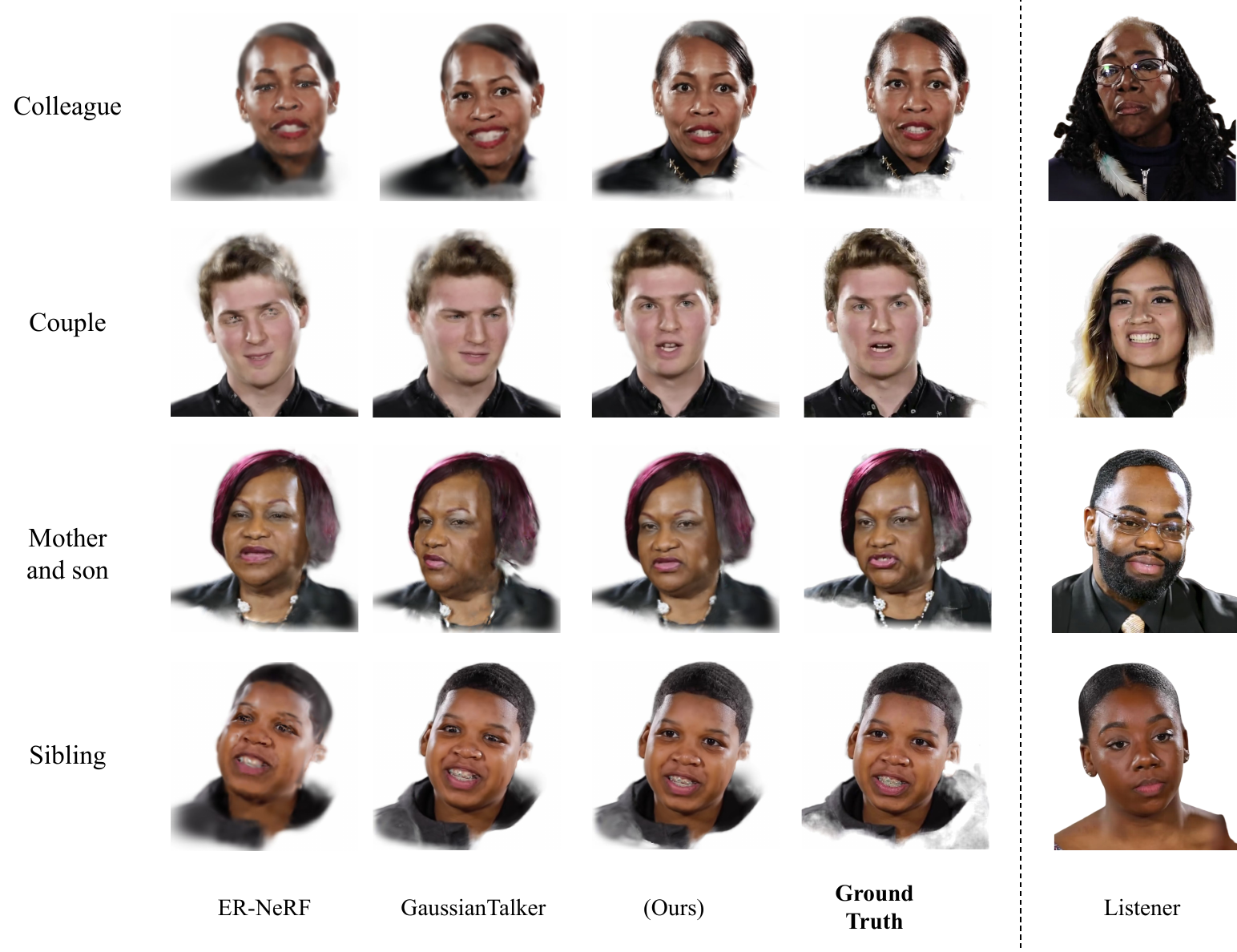}
    \caption{Visualization comparison with the talking head as \textbf{the speaker}. We selected four representative social relationships: superior–subordinate colleagues, couple, mother and son, and siblings.}
    \label{fig:vis_speaker}
\end{figure*}

\begin{figure*}[t]
    \centering
    \includegraphics[width=1.0\textwidth]{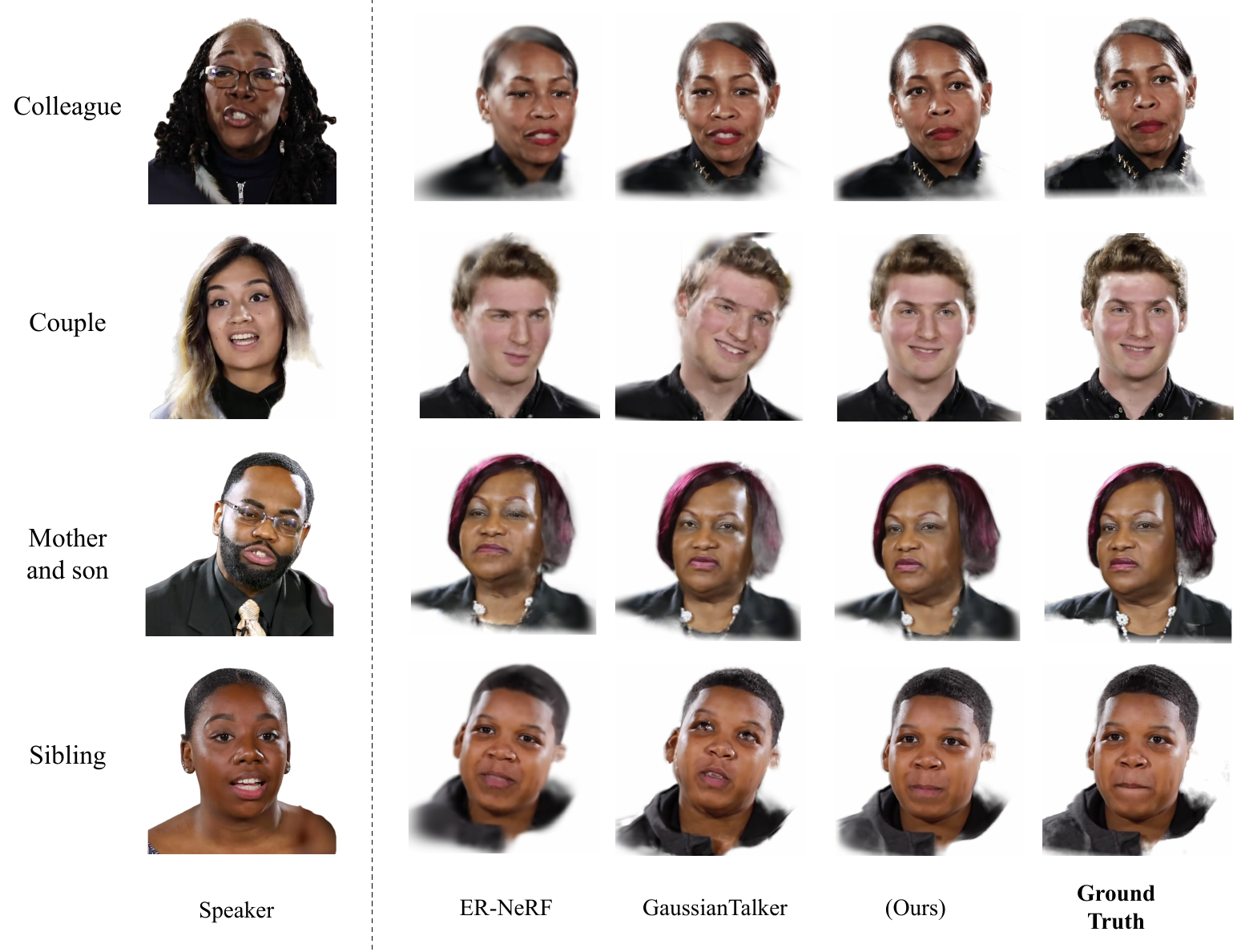}
    \caption{Visualization comparison with the talking head as \textbf{the listener}.}
    \label{fig:vis_listener}
\end{figure*}

\subsection{Qualitative Evaluation}

\subsubsection{Socially-aware user study} \label{sec:user_study}

To further evaluate the rendering quality of talking head generation, we conducted a socially-aware user study comparing RSATalker against three representative baselines: ER-NeRF, SyncTalk, and GaussianTalker. 
Twenty-four participants (graduate students and researchers with prior experience in computer vision and human–computer interaction; 10 male, 11 female, 2 non-binary, and 1 undisclosed, all of whom provided informed consent) each viewed four videos generated from the same input speech and provided continuous ratings (0–100) on nine sub-criteria grouped into three high-level dimensions:
\textbf{Realism} (lip--speech synchronization, expression naturalness, visual photorealism), \textbf{Fluency} (motion smoothness, temporal consistency, long-term stability), and \textbf{Social Relationship Accuracy, SRA} (role consistency, emotional plausibility, interaction dynamics). Dimension scores were computed by averaging the three sub-criteria, and the overall score for each method was computed as
\begin{equation}
S_{\text{total}} = 0.25\,S_{\text{Realism}} + 0.25\,S_{\text{Fluency}} + 0.5\,S_{\text{SRA}}
\end{equation}
reflecting the design choice to emphasize social relationship fidelity in multi-turn conversation.

Tab.~\ref{tab:user_study_full} lists the averaged sub-criterion and total scores. RSATalker achieves the highest scores across every reported sub-criterion and dimension: overall weighted totals are RSATalker = \textbf{86.1}, GaussianTalker = 67.8, SyncTalk = 60.7, and ER-NeRF = 50.2. In relative terms, RSATalker improves the weighted total by \(26.9\%\) over GaussianTalker, \(41.9\%\) over SyncTalk, and \(71.5\%\) over ER-NeRF, indicating a substantial perceptual advantage when social relationship modeling is required.

\paragraph{Dimension-wise findings.}

\textbf{Realism.} RSATalker’s realism average is 88.4 (Lip Sync = 90.7, Expression Naturalness = 86.4, Visual Realism = 88.2), which is \( 7.6\%\) higher than GaussianTalker (81.8) and substantially higher than SyncTalk (77.2) and ER-NeRF (62.0). 
These improvements reflect two factors: (1) accurate audio–visual alignment that reduces micro-mismatches in expressive turns (RSATalker lip sync = 90.7 vs. SyncTalk = 85.2), and (2) Gaussian offset corrections that preserve high-frequency facial details.

\textbf{Fluency.} On temporal metrics, RSATalker averages 86.7 versus GaussianTalker 73.1 and SyncTalk 68.1, an improvement of \( 18.7\%\) over GaussianTalker. The gains are most visible in motion smoothness and long-term stability (RSATalker long-term stability = 85.6 vs. GaussianTalker = 70.9), indicating the effectiveness of speaker-listener motion generator in preventing jitter and motion error across timesteps.

\textbf{Social Relationship Accuracy (SRA).} The largest relative gains are observed in SRA: RSATalker = 84.5 vs. GaussianTalker = 58.1, a relative increase of \( 45.4\%\). 
Sub-criterion improvements are particularly strong for role consistency (84.7 vs. 55.4, \( 52.9\%\) relative gain), emotional plausibility (82.3 vs. 58.7, \( 40.1\%\)), and interaction dynamics (86.5 vs. 60.3, \( 43.5\%\)). These results directly support the hypothesis that conditioning on social relationship embeddings produces observable, interpretable behavioral cues (role-appropriate head poses, emotionally congruent micro-expressions, turn-taking / backchannel patterns) that human raters use to infer relationship and intent.

\paragraph{Sub-criterion analysis and qualitative corroboration.}
Inspecting the sub-criteria in Tab.~\ref{tab:user_study_full} highlights complementary strengths and weaknesses of the baselines. SyncTalk is strong in framewise lip alignment (85.2) but loses ground on temporal consistency (65.7) due to occasional head-pose jitter. GaussianTalker attains high per-frame visual fidelity (Visual Realism = 86.2) via 3D Gaussian splatting but shows lower SRA (58.1) because it does not exploit interlocutor cues; this often results in listener outputs that are visually sharp but socially incongruent (e.g., passive or mismatched listener responses). ER-NeRF tends toward over-smoothing (lower realism and fluency scores) and produces artifacts in high-frequency facial regions (mouth corners, eye contours), which reduces expressiveness and perceived authenticity.

Qualitative inspection of representative clips (Fig.~\ref{fig:vis_speaker} and Fig.~\ref{fig:vis_listener}) confirms these measurements: RSATalker clips contain clearer nodding, timely gaze shifts, and expression micro-variations that align with conversational roles, whereas baselines frequently fail to manifest these socially diagnostic signals.

\paragraph{Limitations.}
Although RSATalker outperforms baselines on average, several limitations remain. 
First, extreme head rotations and heavy occlusions still reduce both visual fidelity and SRA, as key facial cues become unavailable. 
Second, in very long conversations where the model must retain information spanning many turns, we observed occasional attenuation of early-turn social cues, which can slightly reduce role-consistency over extended sequences. 
Third, the present user study was conducted with a domain-expert participant pool (CV / HCI researchers), which is appropriate for sensitivity to subtle cues but may not reflect judgments of lay viewers; future user studies should include broader, more diverse populations to assess ecological validity.

\paragraph{Conclusion of the study.}
In summary, both quantitative scores and qualitative inspection indicate that implicitly modeling social relationships substantially improves the perceived realism, temporal fluency, and—most notably—social relationship accuracy of generated talking heads. The largest and most meaningful gains are in SRA, corroborating our claim that social conditioning is essential for believable multi-turn conversational avatars. 

\subsubsection{Comparison on visualization}

We provide qualitative comparisons with ER-NeRF~\cite{li2023efficient} and GaussianTalker~\cite{cho2024gaussiantalker}. Fig.~\ref{fig:vis_speaker} and Fig.~\ref{fig:vis_listener} present representative frames extracted from multi-turn conversation to highlight the differences in both speaker and listener scenarios.  

In Fig.~\ref{fig:vis_speaker}, we visualize the case where the talking head serves as the speaker. We choose four representative social relationships: superior-subordinate colleagues, couple, mother and son, and siblings.
Our method produces head and facial motions that are not only realistic in appearance but also consistent with the social relationship between two speakers. In contrast, ER-NeRF often generates overly smoothed expressions that fail to capture the high-frequency dynamics of facial muscles, leading to a lack of expressiveness. It also tends to produce minor rendering artifacts, especially around challenging regions such as the mouth corners and eye contours. 
GaussianTalker leverages 3DGS to maintain high rendering fidelity, successfully preserving fine-scale facial texture. However, its motion modeling occasionally suffers from drifting facial landmarks, resulting in misalignment between expressions and the underlying mesh, which weakens the sense of embodiment and continuity. For example, in the superior–subordinate colleague relationship, GaussianTalker displayed an incorrect ``happy'' facial expression instead of the intended ``serious'' one.

Fig.~\ref{fig:vis_listener} illustrates the case where the talking head assumes the listener role. The results clearly show that ER-NeRF and GaussianTalker both fail to generate appropriate facial responses. This is because these methods are fundamentally designed for speaker-only scenarios: they condition generation solely on the speaker’s audio input, while ignoring the multi-modal signals coming from the conversational partner. As a result, they cannot produce socially-aware listener behaviors such as attentive gaze shifts, responsive nods, or subtle backchanneling cues. The lack of conditioning on conversational context leads to visual outputs that appear passive, unengaged, or even unnatural when placed in multi-turn conversation.  

By contrast, our method explicitly incorporates both the speaker’s and the listener’s multi-modal cues, together with social relationship conditioning, which enables the model to generate temporally coherent and socially consistent behaviors across turns. The avatars produced by our approach not only maintain sharp facial details and smooth motion trajectories, but also capture subtle non-verbal behaviors---such as nodding, eyebrow raises, and gaze adjustments---that reflect the interpersonal dynamics of the conversation. These cues play a critical role in signaling attention, empathy, and social alignment, thereby enhancing the realism and credibility of the generated talking heads.

In summary, our method achieves higher visual fidelity and stronger social awareness than baselines, offering a more complete and context-sensitive simulation of conversational interactions.

\subsection{Ablation Studies}

To validate the effectiveness of the socially-aware module and the three-stage training strategy, we perform ablation experiments on the RSATalker dataset. We evaluate several variants of our model: RSATalker without the socially-aware module (w/o SAM), which removes the social embedding input and Gaussian offset corrections; RSATalker trained without the cold-start pretraining of the speaker-listener motion generator (w/o Stage 1); RSATalker trained without the cold-start pretraining of the realistic head avatar renderer (w/o Stage 2); and the full RSATalker model with all modules and three-stage training. 
User studies were conducted following the same protocol as in Sec.~\ref{sec:user_study}, with each variant evaluated on Realism, Fluency, and Social Relationship Accuracy (SRA) using a 0–100 rating scale.

\begin{table}[h]
    \centering
    \caption{Ablation study results (0--100 scale). Removing the socially-aware module or skipping pretraining stages leads to performance drops, particularly in SRA.}
    \begin{tabular}{l|ccc|c}
    \toprule
    Method & Realism & Fluency & SRA & Total \\
    \midrule
    w/o SAM & 85.5 & 78.1 & 71.3 & 76.6 \\
    w/o Stage 1 & 81.3 & 70.9 & 61.3 & 68.7 \\
    w/o Stage 2 & 72.0 & 75.9 & 77.5 & 75.7 \\
    Full pipeline & \textbf{88.4} & \textbf{86.7} & \textbf{84.5} & \textbf{86.1} \\
    \bottomrule
    \end{tabular}
    \label{tab:ablation}
\end{table}

As shown in Tab.~\ref{tab:ablation}, removing the socially-aware module results in a significant drop in SRA, highlighting its importance for capturing interpersonal relationships. 
Skipping the cold-start pretraining of either the motion generator or the realistic head avatar renderer leads to reduced realism and fluency, demonstrating that proper initialization is crucial for stable training. Overall, the full RSATalker achieves the highest scores across all metrics, confirming the effectiveness of integrating the socially-aware module with the three-stage training strategy.

\begin{table}[h]
    \centering
    \caption{Image quality metrics (L1, PSNR, SSIM, LPIPS) for ablation variants. The full RSATalker consistently achieves better image fidelity and perceptual quality.}
    \begin{tabular}{l|cccc}
    \toprule
    Method & L1 $\downarrow$ & PSNR $\uparrow$ & SSIM $\uparrow$ & LPIPS $\downarrow$ \\
    \midrule
    w/o SAM & 0.0205 & 21.4786 & 0.9348 & 0.0613 \\
    w/o Stage 1 & 0.0279 & 20.5629 & 0.9088 & 0.0929 \\
    w/o Stage 2 & 0.0210 & 21.6408 & 0.9140 & 0.0668 \\
    Full pipeline & \textbf{0.0198} & \textbf{22.9979} & \textbf{0.9377} & \textbf{0.0560} \\
    \bottomrule
    \end{tabular}
    \label{tab:ablation_image}
\end{table}

These results indicate that both the socially-aware module and the staged pretraining strategy not only improve social relationship modeling but also enhance the image fidelity and perceptual quality of generated talking heads. The combination of socially-aware user study and traditional image metrics provides a comprehensive understanding of each component's contribution to RSATalker.

\section{Conclusion}


In this work, we present RSATalker, a novel framework for generating realistic and socially-aware talking heads in multi-turn speaker–listener interaction. By integrating 3D Gaussian Splatting with a socially-aware module that models interpersonal relationships, our method effectively produces facial expressions, head movements, and interaction dynamics that are consistent with the underlying social context. The framework is further strengthened by a carefully designed three-stage training strategy and the newly curated RSATalker dataset, ensuring both high fidelity and robust performance across diverse scenarios. Extensive experiments and user studies demonstrate that our method outperforms state-of-the-art baselines in realism, fluency, and social relationship accuracy. By providing a new benchmark for socially-conditioned talking head generation, RSATalker represents a significant step forward in immersive virtual reality applications.

{
    \small
    \bibliographystyle{ieeenat_fullname}
    \bibliography{main}
}


\end{document}